\def\year{2022}\relax
\documentclass[letterpaper]{article} 
\usepackage{aaai22}  
\usepackage{times}  
\usepackage{helvet}  
\usepackage{courier}  
\usepackage[hyphens]{url}  
\usepackage{graphicx} 
\usepackage{natbib}  
\usepackage{caption} 
\DeclareCaptionStyle{ruled}{labelfont=normalfont,labelsep=colon,strut=off} 
\usepackage{amsmath}
\usepackage{multicol}
\usepackage{algorithm}
\usepackage{algorithmic}
\usepackage{adjustbox}
\usepackage{newfloat}
\usepackage{listings}
\floatstyle{ruled}
\newfloat{listing}{tb}{lst}{}
\floatname{listing}{Listing}
\nocopyright
\title{CL-NERIL: A Cross-Lingual Model for NER in Indian Languages}
\author{
    Akshara Prabhakar\textsuperscript{\rm 1},
    Gouri Sankar Majumder\textsuperscript{\rm 2},
    Ashish Anand\textsuperscript{\rm 2}
}
\affiliations{
    \textsuperscript{\rm 1}National Institute of Technology Karnataka, Surathkal \\
    \textsuperscript{\rm 2}Indian Institute of Technology Guwahati \\

    akshblr555@gmail.com,
    gourisankar@iitg.ac.in,
    anand.ashish@iitg.ac.in
}

\usepackage{bibentry}

\begin{document}
\maketitle

\begin{abstract}
Developing Named Entity Recognition (NER) systems for Indian languages has been a long-standing challenge, mainly owing to the requirement of a large amount of annotated clean training instances. This paper proposes an end-to-end framework for NER for Indian languages in a low-resource setting by exploiting parallel corpora of English and Indian languages and an English NER dataset. The proposed framework includes an annotation projection method that combines word alignment score and NER tag prediction confidence score on source language (English) data to generate weakly labeled data in a target Indian language. We employ a variant of the Teacher-Student model and optimize it jointly on the pseudo labels of the Teacher model and predictions on the generated weakly labeled data. We also present manually annotated test sets for three Indian languages: \textit{Hindi}, \textit{Bengali}, and \textit{Gujarati}. We evaluate the performance of the proposed framework on the test sets of the three Indian languages. Empirical results show a minimum 10\% performance improvement compared to the zero-shot transfer learning model on all languages. This indicates that weakly labeled data generated using the proposed annotation projection method in target Indian languages can complement well-annotated source language data to enhance performance. Our code is publicly available at \lstinline|https://github.com/aksh555/CL-NERIL|.
\end{abstract}

\section{Introduction}
Named Entity Recognition (NER), a fundamental task in natural language processing, deals with detecting named entities and their classification into certain predefined categories. However, the performance of NER models depends on the amount and quality of annotated data available. Indian languages, for one, lack such  accurate entity annotated corpora to build and benchmark NER systems. Cross-lingual NER attempts to address this challenge by transferring knowledge from a high-resource source language having abundant labeled entities to a low-resource target language having few or no labels.
The projection-based approaches \cite{mayhew-etal-2017-cheap, xie-etal-2018-neural} applying word translations perform poorly primarily due to syntactic word order differences between English and Indian languages.
Problems arising due to word order differences can be mitigated using word alignment method. This work uses multilingual embedding-based word alignment method to propose an annotation projection method to generate weakly labeled data in the target language. The annotation projection method is integrated into an end-to-end \underline{c}ross-\underline{l}ingual transfer learning framework for \underline{NER} in \underline{I}ndian \underline{l}anguages, referred to as CL-NERIL. Primary contributions of this paper are : (1) An end-to-end framework, CL-NERIL, with a novel annotation projection method and joint optimization, (2) Clean test sets of at least 1000 sentences for the three Indian languages, (3) CL-NERIL obtains at least $10\%$ improvement compared to zero-shot baseline models. 

\section{Proposed Approach}
\textbf{Weakly Labeled Data Generation:} We propose an annotation-projection-based approach to create weakly labeled data from a large bi-lingual parallel corpus \cite{ramesh2021samanantar}. 
Let $\{(S_{i}, T_{i})\}_{i=1}^{N}$ having $N$ sentences be a parallel corpus of English and a low-resource language, where $ (S_{i}, T_{i})$ represents a pair of sentences which are translations of each other. 
We first generate the named entity tag-sequence $ y_{i}^{s} $ 
for the source (English) sentence $ S_{i} $ using Flair NER trained on CoNLL dataset \cite{tjong-kim-sang-de-meulder-2003-introduction} and store the confidence score of each detected entity as \textit{NER score}. Next, we determine word and subword alignments between $ S_{i}$ and $ T_{i} $. We use XLM-R \cite{conneau-etal-2020-unsupervised} for word-level alignment. 
We align two words that are the most similar in terms of normalized cosine similarity. Ties are resolved by giving preference to aligned smaller index word in the sentence.
This is a local optimal solution, and only mutual alignments are identified, resulting in many entity words getting missed. Since our focus is to project the entity tags to the low resource language, we try to further align the missed entity words (if any) by using the Match algorithm \cite{jalili-sabet-etal-2020-simalign}. Here, we utilize the subword-level alignments with mBERT \cite{devlin-etal-2019-bert} embeddings. This process ensures a near alignment for all entity words identified in the source side to a word(s) on the target side. The normalized cosine similarity that we obtain via these alignments for every word is the \textit{alignment score}. In case a word gets aligned to many words, we assign the same tag to all words, and to resolve multiple tag matches, we take the one giving the highest score.   
Then, we propagate the source tag label $y_{i}^{s}$ to target sentence according to the alignments to get $y_{i}^{t}$. Finally, we select the top 40\% (empirically chosen value) sentences based on the sentence score to filter the good quality sentences as weakly labeled data.
Sentence score for a sentence with $k$ number of entities is calculated as
\begin{gather*}
\label{eq:sentscore}
    score = \frac{1}{k} \sum_{i=1}^{k} \log(\text{alignment score} * \text{NER score})
\end{gather*}

\textbf{Teacher-Student Model: }
We employ Teacher-Student model to transfer knowledge from source to target language. Firstly, for the Teacher model, we fine-tune mBERT by training it on CoNLL English data. The Student model is initialized with pre-trained mBERT layers.
Next, our weakly labeled sentences are passed to both the Teacher model and Student model to generate pseudo labels $ y^{teach}, y^{stud} $ respectively, i.e., probability distributions over the tag set for each word $ w_{i} $ in the target sentence.
Finally, the Student model is trained by optimizing the mean squared error (MSE) loss between the two predicted distributions $ y^{teach}, y^{stud} $ (considering the Teacher model's predicted labels as soft), and negative log-likelihood (NLL) of the generated weak label of each word ($y^t$) and the predicted labels ($ y^{stud} $). Model parameter details and hyperparameter tuning are discussed in the supplementary material.
\begin{gather*}
\label{}
    Loss = \frac{1}{N} \sum_{i=1}^{N} \{  MSE(y_{i}^{teach}, y_{i}^{stud}) + NLL(y_{i}^{stud}, y_{i}^{t}) \} 
\end{gather*}

\section{Results \& Analysis}
We evaluate CL-NERIL on manually annotated test sets having at least 1000 sentences for three languages -- \textit{Bengali (bn)}, \textit{Gujarati (gu)}, and \textit{Hindi (hi)}.
We compared our model with three state-of-the-art models for cross-lingual NER. Table \ref{tab:results} summarizes the results. \citet{wu-dredze-2019-beto} used mBERT model for zero-shot transfer (which we use as the Teacher model in our approach) to train NER on English and showed cross-lingual transferability of the model on different target languages. 
We compared the self-learning model, BOND \cite{liang2020bond}, which is one of the strongest baselines on the weakly labeled data (generated by our projection approach). Teacher Student model \cite{wu2020single-} uses knowledge distillation method \cite{sanh2020distilbert} to transfer knowledge of the NER model trained on English to the target language and showed improvement over the zero-shot transfer. 
To assess the quality of our generated weak labels by annotation-projection approach, we trained BERT based monolingual NER model on the weakly labeled data. Its performance corroborated the quality of our generated weakly labeled data.
Our model, CL-NERIL, due to joint optimization of the MSE loss with teacher prediction and NLL loss on weak labels, does a better job in cross-lingual transfer and showed significant improvement over the Teacher Student model on all languages, and even over BOND for two languages.  

\begin{table}[h!]
    \centering
    \begin{adjustbox}{max width=0.4\textwidth}
    \def\arraystretch{1.2}%
    \begin{tabular}{|l|c|c|c|}
        \hline
        \textbf{Model} & \textbf{bn} & \textbf{gu} & \textbf{hi} \\
        \hline
        Zero-shot \cite{wu-dredze-2019-beto} &	50.67 &	56.59 &	70.28 \\
        \hline
        BOND \cite{liang2020bond} &	64.19 &	79.28 &	\textbf{80.14}  \\
        \hline
        Teacher-Student \cite{wu2020single-} & 	53.79 & 	58.80 & 	72.37  \\
        \hline
        \hline
        \textbf{Monolingual on Weak Data (Ours)} & 71.21 & 80.39 & 79.80 \\
        \hline
        \textbf{CL-NERIL (Ours)} &	\textbf{73.34} &	\textbf{80.73 }&	79.69 \\
        \hline
        \end{tabular}
        \end{adjustbox}
    \caption{Benchmarking CL-NERIL against State-of-the-art.}
    \label{tab:results}
\end{table}

\section{Conclusion \& Future Work}
In this work, CL-NERIL, an end-to-end cross-lingual model for NER task on Indian languages in low-resource settings was presented. It leverages the easily procurable weakly labeled data in the target language to complement the gold standard data in the source language to enhance the performance on target languages. We observe that there are two sources of noise in this approach -- from alignment and NER on the source side. We are currently working 
on a novel noise-aware loss function to take care of these noise sources.

\bibliography{aaai22}

\end{document}